\PassOptionsToPackage{unicode}{hyperref}
\PassOptionsToPackage{hyphens}{url}
\PassOptionsToPackage{dvipsnames,svgnames,x11names}{xcolor}
\documentclass[
  10pt,
  twocolumn]{article}
\usepackage{xcolor}
\usepackage[margin=0.7in]{geometry}
\usepackage{amsmath,amssymb}
\usepackage{iftex}
\ifPDFTeX
  \usepackage[T1]{fontenc}
  \usepackage[utf8]{inputenc}
  \usepackage{textcomp} 
\else 
  \usepackage{unicode-math} 
  \defaultfontfeatures{Scale=MatchLowercase}
  \defaultfontfeatures[\rmfamily]{Ligatures=TeX,Scale=1}
\fi
\usepackage{lmodern}
\ifPDFTeX\else
\fi
\IfFileExists{upquote.sty}{\usepackage{upquote}}{}
\IfFileExists{microtype.sty}{
  \usepackage[]{microtype}
  \UseMicrotypeSet[protrusion]{basicmath} 
}{}
\usepackage{graphicx}
\makeatletter
\newsavebox\pandoc@box
\newcommand*\pandocbounded[1]{
  \sbox\pandoc@box{#1}%
  \Gscale@div\@tempa{\textheight}{\dimexpr\ht\pandoc@box+\dp\pandoc@box\relax}%
  \Gscale@div\@tempb{\linewidth}{\wd\pandoc@box}%
  \ifdim\@tempb\p@<\@tempa\p@\let\@tempa\@tempb\fi
  \ifdim\@tempa\p@<\p@\scalebox{\@tempa}{\usebox\pandoc@box}%
  \else\usebox{\pandoc@box}%
  \fi%
}
\def\fps@figure{htbp}
\makeatother
\providecommand{\tightlist}{%
  \setlength{\itemsep}{0pt}\setlength{\parskip}{0pt}}
\usepackage{microtype}
\usepackage{xurl}
\usepackage{float}
\usepackage{fancyhdr}
\AtBeginDocument{\hypersetup{colorlinks=true,linkcolor=black,urlcolor=blue,citecolor=black}}
\usepackage{bookmark}
\IfFileExists{xurl.sty}{\usepackage{xurl}}{} 
\hypersetup{
  pdftitle={Keep the Process{,} Change the Supplier: Runtime Governance for Tool-Using AI Agents},
  pdfauthor={Deepak Waghmare (Two Much Labs); Shaad Sufi (Two Much Labs)},
  colorlinks=true,
  linkcolor={Maroon},
  filecolor={Maroon},
  citecolor={Blue},
  urlcolor={Blue},
  pdfcreator={LaTeX via pandoc}}

\title{Keep the Process, Change the Supplier: Runtime Governance for
Tool-Using AI Agents}
\author{Deepak Waghmare\\{\normalsize Two Much Labs}\\{\small\href{mailto:deepak@twomuchlabs.com}{deepak@twomuchlabs.com}}\and Shaad Sufi\\{\normalsize Two Much Labs}\\{\small\href{mailto:shaad@twomuchlabs.com}{shaad@twomuchlabs.com}}}
\date{19 November 2026}

\begin{document}
\maketitle

\begin{abstract}

Tool-using AI agents propose actions whose authorization and external
effects cannot be inferred from generated text alone. We study an
execution boundary that separates agent proposals from
business-controlled authority, context disclosure, and supplier
selection. The Open Process Control Plane (OPCP) mediates capability
calls and records uncertain outcomes when a supplier may have acted
without returning an acceptable reply. We evaluate this runtime with
scripted requests and synthetic suppliers: 2,400 fault episodes across
four policies, 60 process-termination/restart episodes, 19,000 timed
calls across 19 configurations, five targeted mutations over 43
fixtures, and 160 substitution records. No language model or autonomous
agent is evaluated. OPCP and a conservative composition agree on all
measured fault outcomes. Both recover after definite refusal without
cross-supplier duplication; retry-all duplicates effects in every tested
lost-reply-after-commit case. Persistent runtime state prevents
cross-supplier replay in the restart experiments, while same-supplier
duplicate prevention also requires supplier idempotency. At 128
suppliers, best-score routing has a median latency of 1.411 ms versus
0.261 ms for pinned routing. All five selected mutations are detected,
and record enumeration exposes acceptance of an empty regression set.
These results characterize an execution mechanism relevant to agent
governance, including its costs and assumption boundaries. They do not
establish improved agent task success, prompt-injection resistance, or
real-vendor model portability.

\end{abstract}

\section{1 Introduction}\label{introduction}

A tool-using AI agent can propose an action, receive an ambiguous
response, and propose another action to complete its task. The execution
layer must decide whether the action is authorized, what information the
tool may receive, and whether retrying through another provider could
repeat an external effect. These decisions remain necessary even when
the agent's proposed action is syntactically valid. A generated account
of success is also distinct from evidence that the intended effect
occurred once.

AgentDojo evaluates agents interacting with tools and untrusted data,
making prompt injection and task completion explicit evaluation targets
{[}29{]}. ToolEmu uses a language-model-emulated sandbox to identify
risky tool-use behavior {[}30{]}. We address a complementary layer: the
runtime that mediates a proposed capability call after it leaves the
agent. Our experiments isolate that layer with scripted requests. They
do not execute either benchmark, evaluate a language model, or establish
resistance to prompt injection.

The research question is: \textbf{which execution decisions can be kept
outside the agent's generated output, and what happens when a tool
supplier fails or changes?} OPCP places input projection, authority
checks, hard supplier eligibility, selection, and outcome records at an
explicit boundary. The proposal policy can change without granting it
control over those loaded rules. This is a design for integrating agents
with external capabilities; its mechanisms also apply to conventional
workflows. No new reasoning algorithm, learned policy, consensus
protocol, or transaction protocol is claimed.

Supplier substitution makes this separation operationally relevant.
Switching costs and lock-in are established concerns {[}10, 11{]}, and
prompts may require adaptation across models {[}12{]}. ECB Banking
Supervision reported 82\% of externally outsourced critical functions
difficult or impossible to substitute in its 2024 analysis of year-end
2023 data {[}8{]}. DORA, APRA, MAS, Basel, and FSB publications address
related third-party risks with different scopes {[}2-7{]}; the ESAs
reported 3,383 major ICT incidents in 2025, 29\% attributed to
third-party failure {[}9{]}. These observations motivate maintaining an
owned execution boundary; they are not evidence of OPCP effectiveness or
regulatory compliance.

The contributions are:

\begin{enumerate}
\def\labelenumi{\arabic{enumi}.}
\tightlist
\item
  An agent-integration model that separates proposed actions from
  delegated authority and distinguishes uncertain execution from
  definite failure.
\item
  A controlled empirical characterization of the capability runtime,
  including functional controls, process termination, targeted
  mutations, scaling, and evidence-record enumeration.
\item
  An account of the resulting integration obligations: complete
  mediation, host-controlled identity and retry keys, supplier
  idempotency, and task-level evaluation beyond accepted tool calls.
\end{enumerate}

Beldi and Boki provide methodological precedents for measuring stateful
execution and recovery {[}27, 28{]}. RouteLLM evaluates model routing
against cost-quality objectives {[}20{]}. Their workloads and guarantees
differ from ours; published latency numbers are not treated as matched
comparisons. The present contribution is an architectural and mechanism
study relevant to AI-agent execution. End-to-end agent benefit and
reduced organizational switching cost remain hypotheses.

\section{2 Model and implementation}\label{model-and-implementation}

\subsection{2.1 Separating proposals from
authority}\label{separating-proposals-from-authority}

Let an agent policy \(\pi_\theta\) propose a capability request \(q_t\)
from interaction history \(h_t\). A host passes the proposal, an
authenticated principal, and an operation key \(k_t\) to an execution
mediator \(\Gamma\):

\[q_t=\pi_\theta(h_t),\qquad (o_t,z_{t+1})=\Gamma(q_t,k_t;D,z_t).\]

Here \(D\) contains the business-controlled configuration, \(z_t\) is
runtime replay/pending state, and \(o_t\) is the structured response.
This notation describes an integration boundary, not a new agent
algorithm or a formal verification result. The experiments replace
\(\pi_\theta\) with a script and measure the behavior of \(\Gamma\).
They therefore identify execution-layer properties without measuring how
a model chooses actions or interprets the resulting observations.

The design objective is that a new supplier invocation requires valid
leased input, an allowed authority decision, and an eligible supplier.
These conditions are checks on declared policies, not a semantic proof
that the request advances the user's goal. For example, an erroneous but
authorized ledger entry may satisfy the schema and still be a bad
action. Response validation occurs after invocation; rejecting a
malformed reply cannot undo an effect already committed by the supplier.

An integrating host must establish identity and authority outside
model-controlled arguments. It must also preserve operation identity
across retries: the implementation scopes replay memory by capability,
principal, and idempotency key. A model that requests the same
real-world action under a new key can cause a second effect. OPCP does
not infer semantic equivalence between differently keyed requests. The
study evaluates stable-key behavior; host enforcement of keys across an
actual agent's planning loop is not tested.

\subsection{2.2 Ownership and supplier
selection}\label{ownership-and-supplier-selection}

Represent a deployment as \(D=(P,C,A,S)\): process definition \(P\),
context-lease definition \(C\), authority policy \(A\), and supplier
configuration \(S\). A supplier-only substitution changes \(S\) while
holding the other artifacts fixed. A capability supplier here implements
a callable tool or service; it need not be the language model that
proposes the request. The experiments do not measure replacement of the
agent's model or preservation of its planning behavior. A
business-quality acceptance condition must be specified separately.
Equality of revision labels is a consistency check; retaining the
artifacts and their content digests is necessary to establish what was
compared.

For a request \(q\), eligibility is a conjunction of hard conditions:

\[E(s,q)=\bigwedge_{j=1}^{7} e_j(s,q).\]

The conditions cover supplier approval, active contract,
conformance/version, data classification, residency, permitted action
class, and health. Selection operates only over eligible suppliers. The
implementation supports pinned, segmented, cascade, weighted, and
best-score routes; best-score is restricted to action classes R0-R1.
Authority is checked before supplier invocation. The gate uses
Reversibility Classes R0-R4 {[}1{]}, with approval, budgets, and
overrides rather than model-generated authorization decisions.

\begin{figure*}[t]
\centering
\pandocbounded{\includegraphics[keepaspectratio,alt={Proposed agent-integration boundary. The host sends an agent or workflow request through business-controlled contracts and policies. The agent integration is architectural; the experiments supply scripted requests. Relevant calls must be mediated, and supplier behavior and asserted metadata remain assumptions.}]{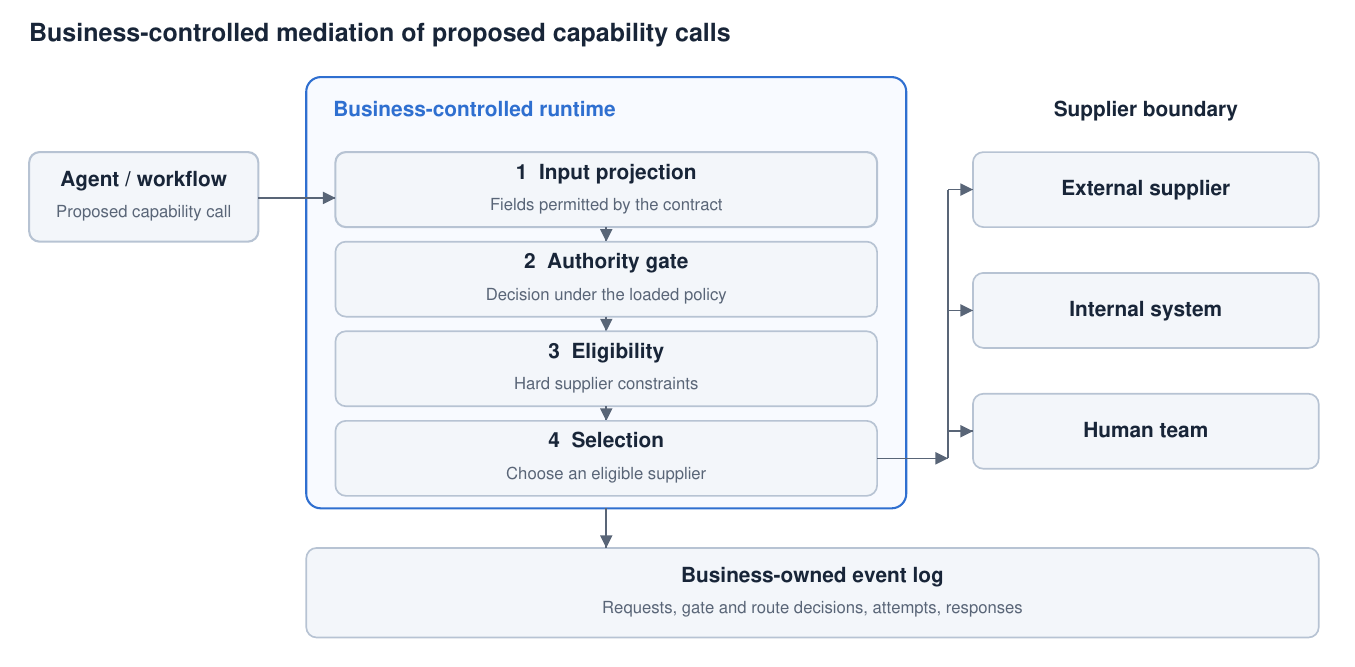}}
\caption{Proposed agent-integration boundary. The host sends an agent or
workflow request through business-controlled contracts and policies. The
agent integration is architectural; the experiments supply scripted
requests. Relevant calls must be mediated, and supplier behavior and
asserted metadata remain assumptions.}
\end{figure*}

A contract declares an archetype, input/output schemas, action class,
and evidence requirements. The seven archetypes are check, extract,
generate, transact, retrieve, notify, and human task. Input projection
removes undeclared fields, including nested fields. Responses are
validated against contract and archetype requirements. The evaluation
covers the example check and transaction contracts and the existing
runtime fixtures; it does not establish completeness of the
seven-archetype taxonomy.

OPCP governs one capability call at a time, not process scheduling or
distributed commit. MCP can supply a tool transport {[}13{]}; a policy
engine such as OPA could supply richer authorization {[}15{]}. BPMN,
OCEL, BIAN, and APQC address process or event representations and
classifications {[}17-19{]}. These are potential integration points, not
integrations validated in this study.

\subsection{2.3 Definite failure, uncertainty, and
replay}\label{definite-failure-uncertainty-and-replay}

A definite failure certifies that the supplier did not perform the
requested effect. An unknown outcome means that the supplier may have
acted but the runtime lacks an acceptable response. For effectful calls,
OPCP stops cross-supplier failover after uncertainty and records the
supplier to reconcile with. A retry using the same key is directed to
that supplier. Completed responses are replayed from runtime state.
Persistent idempotency storage is optional; the default store is in
memory.

The distinction is necessary under the assumed interface. Consider two
executions that yield the same timeout: in one the supplier did nothing;
in the other it committed the effect and lost the reply. If another
supplier performs the command without additional coordination or shared
idempotency, the second execution can duplicate the effect. If the
runtime refuses to continue, the first execution loses immediate
availability. The runtime cannot resolve this ambiguity from the timeout
alone. OPCP chooses conservative handling; the experiments quantify the
consequences for the selected scenarios.

Pinning is not an exactly-once guarantee. A supplier that repeats an
effect on a same-key retry can still duplicate it. Compensation is also
distinct: the transaction contract records a compensation description,
but OPCP is not a saga coordinator {[}16{]}.

\subsection{2.4 Trust assumptions}\label{trust-assumptions}

The trusted boundary comprises the runtime, loaded contracts/policies,
adapters, and deployment infrastructure. The caller's identity and the
truth of supplier approval, classification, and residency metadata are
established elsewhere. A caller with independent supplier credentials
can bypass mediation. Projection cannot revoke disclosed data or prevent
a recipient retaining or inferring information from it.

Runtime persistence and supplier-side idempotency are separate
requirements. The file store is designed for one writer process; shared
multi-process coordination is out of scope. Its fsync option is measured
here, but process termination does not simulate power loss or verify
storage-hardware durability. We explicitly test non-deduplicating
suppliers as an assumption-violating condition rather than quietly
excluding their failures.

\subsection{2.5 Substitution evidence}\label{substitution-evidence}

A substitution record compares revisions of \(P,C,A\), a conformance
flag, and regression outcomes. With case set \(T\), the essential
acceptance condition is:

\[V=I_P\land I_C\land I_A\land K\land G(T),\]

where \(I\) denotes unchanged revisions, \(K\) contract conformance, and
\(G(T)\) exact agreement between candidate and expected outcomes for
every case. The implementation additionally reports a material-exception
gate for expected fail/refer cases. Because \(G\) requires agreement on
every case, this extra gate is logically subsumed by \(G\); it is
diagnostically useful but not an independent sixth barrier.

The current implementation permits \(T=\varnothing\), for which the
universal condition is true. Section 4.5 measures that boundary. A
record's internal consistency does not authenticate its origin, prove
that a supplier was run, or measure contracting and onboarding. A shadow
drill's execution time is not an operational end-to-end exit duration.
The companion exit-readiness scorecard uses a conservative
weakest-measure rule and declared/documented/tested evidence levels; its
thresholds and predictive validity are not evaluated here.

\section{3 Experimental design}\label{experimental-design}

\subsection{3.1 Questions and artifact}\label{questions-and-artifact}

We organize the evaluation around four questions:

\begin{itemize}
\tightlist
\item
  \textbf{RQ1:} Which fault/recovery outcomes follow from
  uncertain-outcome handling and persistence?
\item
  \textbf{RQ2:} Do the existing fixtures detect targeted removal of
  safety mechanisms?
\item
  \textbf{RQ3:} How does local execution cost change with supplier
  count, input size, and persistence?
\item
  \textbf{RQ4:} What does the substitution-record validator accept in a
  bounded input space?
\end{itemize}

The evaluated code is OPCP 0.3.2 at commit \texttt{bb2ceab1}. Runtime,
schemas, and fixtures are unchanged. Experiment scripts, a locally
frozen protocol, per-episode traces, timing rows, environment metadata,
and content hashes accompany the paper. This is not an externally
preregistered or independently executed study. Pilot runs used separate
output directories and were not pooled with retained observations. The
pre-existing development suite passed 332 tests, with one release-only
wheel-import test skipped; that count is a reproducibility check, not a
primary effectiveness result.

The unit of evaluation is a capability-call episode, not an autonomous
task or conversation. Model reasoning, prompt following, tool-choice
quality, and attack success are not measured. The four policies in E1
are execution policies, not competing language-model agents.

All retained experiments ran on one Apple M4 Pro workstation with 12
logical CPUs, 24 GiB RAM, macOS 15.5, and Python 3.14.6. Suppliers are
deterministic local adapters with synthetic inputs. No cloud inference,
external supplier API, customer record, or production process is
involved.

\subsection{3.2 Fault policies and independent
observations}\label{fault-policies-and-independent-observations}

E1 crosses 12 scenario families, 50 payload variants, and four policies:
2,400 episodes. The families include clean completion, definite refusal,
pre-effect timeout, post-effect reply loss, malformed or evidence-free
post-effect replies, authority denial, supplier ineligibility, completed
replay, and retries with/without supplier deduplication. A fresh runtime
and observer ledger are used for each episode. The three replay/retry
families receive two caller requests; the others receive one.

The policies are:

\begin{itemize}
\tightlist
\item
  \textbf{OPCP:} the unchanged runtime.
\item
  \textbf{Conservative composition:} separately written orchestration
  sharing the same contract, gate, validation, and eligibility
  primitives. It distinguishes definite failure from uncertainty and
  retains completed and pending keys. This is a functional control, not
  an independent OPCP implementation or a deployed workflow product.
\item
  \textbf{Retry-all:} the composition retries the next eligible supplier
  after every error or invalid response.
\item
  \textbf{Stop-all:} the composition stops on every error, including
  definite refusal.
\end{itemize}

The latter two are diagnostic controls. We do not present them as
competitive production implementations or claim a performance advantage
over the conservative composition. All policies receive the same input,
authorization decision, and supplier configuration.

An observer-owned ledger records each supplier invocation and committed
effect. It is separate from the runtime's response and log, so an
\texttt{ok} response cannot hide two effects. For key \(k\), let \(n_k\)
be committed effects and \(s_k\) the number of distinct suppliers that
committed. We report \(\mathbf{1}[n_k>1]\) and \(\mathbf{1}[s_k>1]\)
separately, plus whether the final response acknowledges success. These
are counts under deliberately injected faults, not estimates of
real-world incident probabilities. Payload variants exercise repetition;
they do not create 50 independent fault models.

\subsection{3.3 Forced process
termination}\label{forced-process-termination}

E5 uses actual worker processes and terminates the first with
\texttt{os.\_exit(73)} at one of three boundaries: before an effect,
after its commit but before the reply, or after an internal runtime
response but before delivery to the caller. Recovery starts a new
process with preferred supplier order reversed from A/B to B/A.
Synthetic supplier effects are committed to a separate SQLite database,
allowing the parent to inspect the resulting state.

We cross three boundaries, two runtime stores (memory or fsync file),
and supplier deduplication enabled/disabled, with five repetitions per
cell: 60 episodes and 120 process launches. Deduplication is scoped to
each supplier; a record at A does not automatically prevent an effect at
B. This experiment tests three controlled termination windows and
reconfiguration, not arbitrary crash timing, distributed partitions, or
power failure.

\subsection{3.4 Mutation and record
experiments}\label{mutation-and-record-experiments}

E2 executes all 43 runtime fixtures against unmodified code and five
single-mechanism variants: input projection disabled, eligibility
disabled, authority denial bypassed, unknown-outcome protection
disabled, and completed-response memory disabled. The modifications are
applied only inside the experiment process. We preserve all 258
fixture/variant outcomes, distinguishing assertion failures from
unexpected harness errors. This is a selected mutation campaign, not an
exhaustive mutation score or a reproduction of the authors' earlier
review history.

E4 exhaustively enumerates a small product space: eight
revision-equality combinations, two conformance values, and ten case
configurations (empty plus all nine expected/candidate pairs over
pass/fail/refer), producing 160 valid records. We compare acceptance
with an explicitly stricter research criterion that also requires a
nonempty case set. The criterion is not misrepresented as the current
normative rule.

\subsection{3.5 Timing and analysis}\label{timing-and-analysis}

E3 comprises 19 configurations: supplier pools of 1/8/32/128 under
pinned or best-score routing; inputs of 2/16/64/256 ledger lines under
memory or chained-file event logging; and memory, unsynced-file, or
fsynced-file idempotency storage for a two-line transaction. Event
logging remains in memory for the persistence comparison. Each
configuration uses ten fresh-runtime rounds, ten untimed warm-up calls,
and 100 timed calls per round: 19,000 measured and 1,900 warm-up calls.

Configuration order is randomized with seed 20261119. Every call uses a
new idempotency key. Setup and correctness assertions are outside the
timer; full synchronous runtime execution is inside. Timing uses
\texttt{perf\_counter\_ns}. The transaction path includes a
deterministic approval callback. File event logs are normally buffered;
their writes do not imply fsync durability. Garbage collection stays
enabled and caches grow within each block. Dependencies, fault
experiments, and subprocess experiments were not intentionally run
concurrently with retained timing; other workstation activity was
uncontrolled.

The text and artifact tables report pooled median and nearest-rank p95.
Scaling plots show the median of round medians with 95\% percentile
block-bootstrap intervals (2,000 resamples). The bootstrap unit is a
round, not an individual dependent call; rounds still share a process
and host. Intervals therefore describe this run's variability, not
cross-machine generalization. No significance test or population
reliability interval is applied to deterministic fault counts.

\section{4 Results}\label{results}

\subsection{4.1 Fault handling exposes a safety-availability
tradeoff}\label{fault-handling-exposes-a-safety-availability-tradeoff}

OPCP and the conservative composition agree on every measured fault
outcome, including invocation/effect counts, final status, and
uncertainty. Both recover in all 50 definite-refusal variants with one
committed effect; stop-all completes none. For the timeout-before-effect
condition, OPCP completes none immediately, although the observer knows
no effect occurred. Retry-all completes all 50 safely in that particular
condition. The runtime cannot observe the oracle's distinction between
this timeout and a post-commit lost reply.

For timeout after effect, invalid post-effect output, and missing
post-effect evidence, retry-all completes all 50 variants per family but
duplicates effects at two suppliers in all 50. OPCP stops with
uncertainty and produces no cross-supplier duplication. Figure 2 reports
all 12 families, including negative controls and assumption-violating
cases, without averaging away these differences.

\begin{figure*}[t]
\centering
\pandocbounded{\includegraphics[keepaspectratio,alt={Fault-policy outcomes for all 12 scenario families. Each cell is a count out of 50 variants. Left: episodes with more than one committed effect, regardless of supplier. Right: final responses acknowledging success. The conservative composition matches OPCP; acknowledged completion alone is not a safety metric. These deterministic scenarios do not estimate deployment failure rates.}]{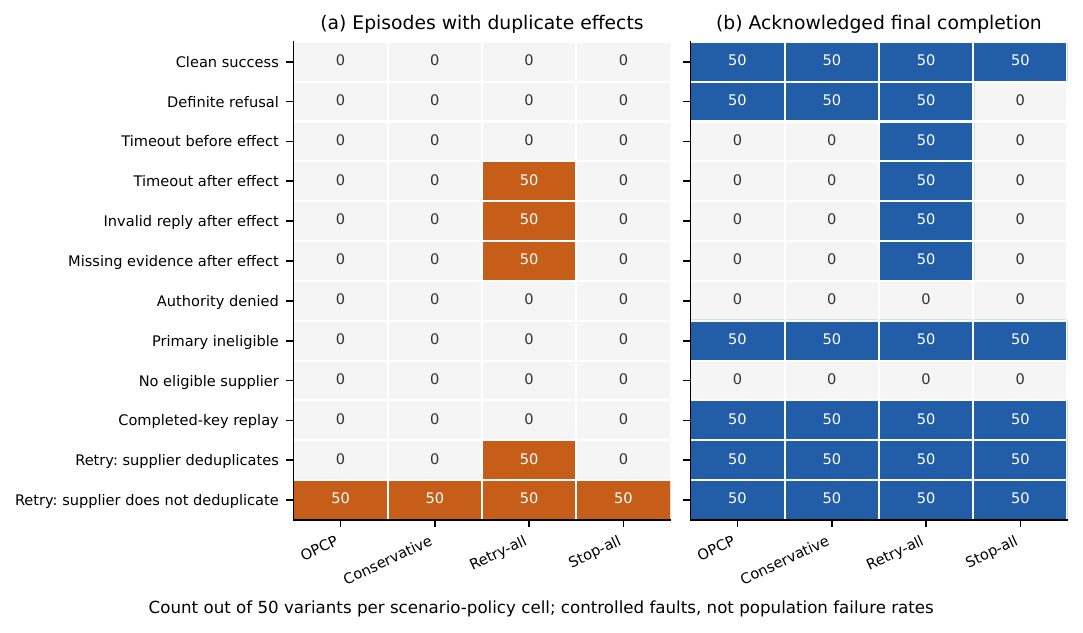}}
\caption{Fault-policy outcomes for all 12 scenario families. Each cell
is a count out of 50 variants. Left: episodes with more than one
committed effect, regardless of supplier. Right: final responses
acknowledging success. The conservative composition matches OPCP;
acknowledged completion alone is not a safety metric. These
deterministic scenarios do not estimate deployment failure rates.}
\end{figure*}

For completed-key replay, all four policies retain one effect per key.
After reply loss, a same-supplier retry with supplier deduplication also
retains one effect under OPCP. Without supplier deduplication, all 50
OPCP variants commit twice at the same supplier. Across the 550 OPCP
variants satisfying the tested supplier assumption, no duplicate effect
is observed; the 50 deliberately nonconforming variants show that
pinning alone is insufficient. Across all 600 OPCP episodes,
cross-supplier duplicates remain zero. The distinction between these two
measures matters more than a single aggregated ``success rate.''

\subsection{4.2 Runtime persistence and supplier idempotency are
complementary}\label{runtime-persistence-and-supplier-idempotency-are-complementary}

All 60 restart episodes eventually return success, but 25 contain
duplicate effects under the tested storage/adapter combinations. This
aggregate is not a reliability estimate; Figure 3 separates the causal
conditions.

After a committed effect with no reply, the memory store duplicates at
the replacement supplier in all five repetitions for each deduplication
setting. Supplier-local deduplication cannot help when recovery chooses
a different supplier. The persistent store instead returns to the
recorded supplier. With supplier deduplication it retains one effect in
all five repetitions; without deduplication it produces a second effect
at the same supplier in all five.

When termination occurs after an internal response, the persistent
completed record allows recovery without reinvocation: no duplicates
occur in either supplier mode. The memory store loses that response and
duplicates after the supplier preference changes. When termination
occurs before any effect, all configurations recover with one effect.
Persistent runtime state therefore addresses cross-supplier replay in
these windows; supplier-side idempotency addresses repetition within the
original supplier. Neither should be reported as a substitute for the
other.

\begin{figure*}[t]
\centering
\pandocbounded{\includegraphics[keepaspectratio,alt={Observed duplicates after forced process termination and recovery with reversed supplier preference. Each bar covers five repetitions. ``Persistent'' is OPCP's fsynced file idempotency store; supplier deduplication is an independent SQLite-backed mechanism. All 60 episodes returned success, illustrating why acknowledgement and effect cardinality must be measured separately.}]{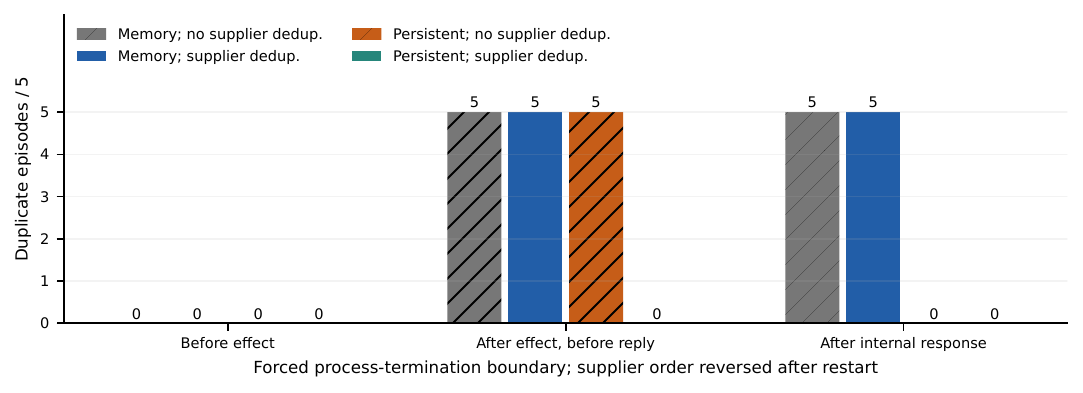}}
\caption{Observed duplicates after forced process termination and
recovery with reversed supplier preference. Each bar covers five
repetitions. ``Persistent'' is OPCP's fsynced file idempotency store;
supplier deduplication is an independent SQLite-backed mechanism. All 60
episodes returned success, illustrating why acknowledgement and effect
cardinality must be measured separately.}
\end{figure*}

\subsection{4.3 Execution cost depends on routing and payload
size}\label{execution-cost-depends-on-routing-and-payload-size}

Pinned routing's median increases from 150.8 microseconds at one
supplier to 261.2 microseconds at 128. Best-score increases from 154.0
to 1,410.8 microseconds. At 128 suppliers, its median is 5.40 times
pinned routing's, with p95 3,074.0 versus 362.5 microseconds. This is a
local algorithm-cost comparison, not a comparison with RouteLLM or an
external model router.

Inspection explains a scaling concern consistent with the observations:
best-score recomputes pool-wide cost and latency maxima for each
candidate, giving quadratic work in pool size before sorting. Pinned
eligibility scanning is linear in the candidate count. The experiment
does not establish a universal asymptotic fit, but it identifies a
concrete optimization opportunity: compute normalization factors once
per pool. No optimized variant is included or claimed as evaluated.

For the transaction input study, increasing ledger lines from 2 to 256
raises memory-logged median latency from 203.3 microseconds to 5,542.4
microseconds. Chained-file logging increases the corresponding medians
from 405.5 to 5,776.7 microseconds. These results invalidate any
unqualified ``sub-millisecond overhead'' claim based on the earlier
small-payload example.

\begin{figure*}[t]
\centering
\pandocbounded{\includegraphics[keepaspectratio,alt={Scaling measurements on one workstation. Markers are medians of ten round medians; error bars are 95\% block-bootstrap intervals over rounds. Left: candidate-pool size with a fixed check input. Right: transaction input size with a single supplier. Both axes use logarithmic scales. Workloads and uncertainty units are described in Section 3.5.}]{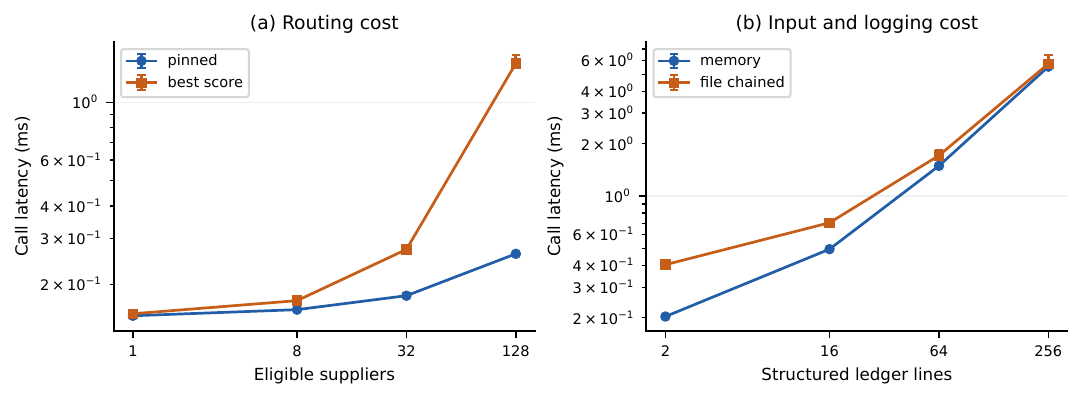}}
\caption{Scaling measurements on one workstation. Markers are medians of
ten round medians; error bars are 95\% block-bootstrap intervals over
rounds. Left: candidate-pool size with a fixed check input. Right:
transaction input size with a single supplier. Both axes use logarithmic
scales. Workloads and uncertainty units are described in Section 3.5.}
\end{figure*}

The persistence comparison uses the same two-line transaction and memory
event log. Median call latency is 200.7 microseconds with in-memory
idempotency, 352.1 with file persistence without fsync, and 418.3 with
fsync. The fsynced configuration costs 2.08 times the in-memory median
here; p95 is 768.8 microseconds. Figure 5 separates the median from the
tail rather than using p95 as an error bar. These are local call
latencies, not durable end-to-end transaction latency or throughput.

\begin{figure}[tbp]
\centering
\pandocbounded{\includegraphics[keepaspectratio,alt={Idempotency persistence cost with a fixed two-line transaction and memory event log. Bars show pooled medians; black diamonds show p95 across 1,000 measured calls per configuration. Diamonds are percentiles, not confidence intervals.}]{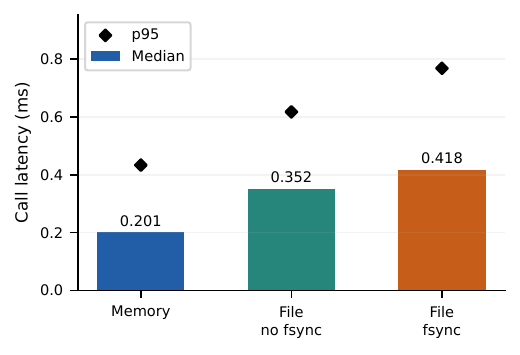}}
\caption{Idempotency persistence cost with a fixed two-line transaction
and memory event log. Bars show pooled medians; black diamonds show p95
across 1,000 measured calls per configuration. Diamonds are percentiles,
not confidence intervals.}
\end{figure}

\subsection{4.4 The selected mutations are detected,
unevenly}\label{the-selected-mutations-are-detected-unevenly}

All 43 runtime fixtures pass on unmodified code. Disabling projection
causes two fixture failures; disabling eligibility causes 15; bypassing
denial causes four; disabling unknown-outcome protection causes three;
and discarding completed-response memory causes two. All five mutants
are detected through assertions, with no unexpected harness errors.
Figure 6 exposes the per-mechanism counts.

\begin{figure}[tbp]
\centering
\pandocbounded{\includegraphics[keepaspectratio,alt={Existing fixtures that detect each single-mechanism removal. All 43 fixtures run for each variant. Every selected mutant is detected; counts describe sensitivity within this manually selected set, not exhaustive coverage or a general security score.}]{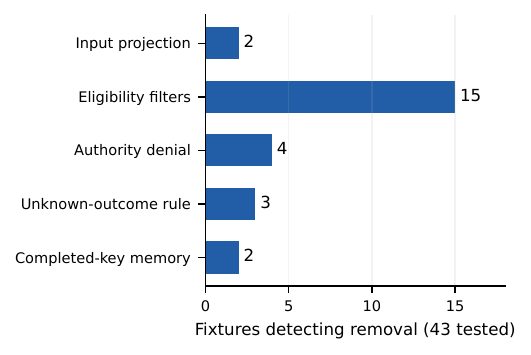}}
\caption{Existing fixtures that detect each single-mechanism removal.
All 43 fixtures run for each variant. Every selected mutant is detected;
counts describe sensitivity within this manually selected set, not
exhaustive coverage or a general security score.}
\end{figure}

The small mutant set cannot establish mutation adequacy. Several
fixtures test the same primitive, and most fixtures are unrelated to any
one mutation. Conversely, removing a mechanism and observing one failing
fixture is more informative than reporting a large unstructured test
count. The artifact retains the complete fixture-by-variant matrix,
including surviving fixtures, so the finding is reproducible and open to
extension.

\subsection{4.5 Record consistency does not establish evidence
sufficiency}\label{record-consistency-does-not-establish-evidence-sufficiency}

All 160 enumerated records are structurally valid. The validator accepts
four. Three are nonempty exact matches with unchanged revisions and
conformance true; the fourth has an empty regression set under the same
conditions. The stricter nonempty criterion accepts three, producing one
disagreement. We report the individual boundary rather than a misleading
general ``99.4\% accuracy'' figure: the sample space was constructed,
and the additional nonempty requirement is a proposed research
criterion.

Adding a nonempty requirement would address this example, but not
provenance or representativeness. One handpicked passing case can still
be inadequate. Revision labels can be equal without an external verifier
possessing the artifacts. Timing fields and conformance flags can be
supplied without corresponding execution. The format supports evidence
exchange and consistency checking; it does not independently certify a
real supplier substitution.

\section{5 Discussion}\label{discussion}

\subsection{5.1 What the comparison
establishes}\label{what-the-comparison-establishes}

The conservative composition's match is an important null comparison.
The evaluated orchestration behavior can be assembled from standard
primitives. OPCP's proposed value is an explicit contract and evidence
boundary shared across substitutions, not exclusive ownership of a
retry-safety mechanism. Proving reduced integration or migration effort
requires a different experiment with real developers, suppliers, and
acceptance criteria.

Beldi and Boki investigate richer stateful execution guarantees and
evaluate application/distributed workloads {[}27, 28{]}. Our local
results do not establish equivalent guarantees or performance. RouteLLM
targets cost-quality allocation between language models {[}20{]}; our
best-score measurements quantify implementation overhead without
measuring answer quality. Contemporary governance proposals also
separate policy from agent execution {[}22-25{]}. OPCP adds a
substitution-oriented representation, but a broad novelty claim would
require a systematic review and stronger interoperability evidence.

FINOS identifies provider availability/tight coupling and authorization
bypass as risks {[}21{]}. The experiments show mechanisms relevant to
these risks under specified assumptions. They do not prove regulatory
compliance or production security. OpenTelemetry could carry execution
observations {[}14{]}, but a validated semantic mapping is future work.

\subsection{5.2 Implications for tool-using
agents}\label{implications-for-tool-using-agents}

The fault results separate three questions an agent controller must not
collapse: was the action permitted, did an effect occur, and does the
caller have acceptable evidence of it? In the post-effect lost-reply
condition, the observer records an effect while OPCP returns an unknown
outcome. Interpreting every failed response as proof of no effect would
invalidate the conservative policy. A host should propagate this
uncertainty and preserve the operation key when a subsequent agent turn
requests a retry. This is an integration requirement inferred from the
tested mechanism, not an observed improvement in model behavior.

Input projection and an authority gate constrain declared interfaces;
they do not identify malicious instructions in tool output, establish
the truth of returned facts, or judge whether a permitted action serves
the user's intent. Complete mediation is necessary: an agent with direct
provider credentials can bypass the boundary. These limits distinguish
our evaluation from the agent-level attack and task analyses in
AgentDojo and ToolEmu {[}29, 30{]}.

An end-to-end follow-up should hold the agent model and tasks fixed
while comparing direct tool access, a competent policy composition, and
OPCP under matched permissions. It should independently record task
completion, forbidden and duplicate effects, unnecessary refusals,
unresolved outcomes, and how the controller handles unknown responses.
Multiple models, held-out tasks, and adversarial tool observations would
test whether the execution boundary improves agent behavior without
merely blocking useful work. None of these outcomes is claimed from the
present scripted study.

\subsection{5.3 Requirements for an operational substitution
claim}\label{requirements-for-an-operational-substitution-claim}

A meaningful next study should compare OPCP with a competent
workflow-plus-policy composition using the same real suppliers and
held-out business cases. It should measure initial adapter/contract
development as well as later migration effort, preserve before/after
content digests, and assess quality and material exceptions separately.
Contracting, onboarding, and technical cutover should be recorded with
explicit start/stop events, including overlap and incomplete exits.

A shadow test cannot establish production cutover time or absence of
customer-visible side effects by declaration alone. The default event
log withholds values and is not itself a replay dataset. Optional hash
chaining also needs an independently retained head to detect tail
truncation; a chain does not authenticate its writer. These boundaries
should be part of the acceptance criteria, not hidden behind the word
``proven.''

\section{6 Threats to validity}\label{threats-to-validity}

\textbf{Construct validity.} Duplicate effects and acknowledgements
measure selected execution properties, not all business harm. A
synthetic ledger does not model settlement, compensation success,
semantic correctness, or a human decision. The non-deduplicating
supplier conditions intentionally break an assumption. They expose a
boundary rather than estimate normal-operation behavior.

\textbf{Internal validity.} Policies share contract, authority, and
eligibility components and the same experiment author, so agreement is
not independent verification. Deterministic faults and fresh-runtime
episodes simplify causal attribution while omitting interacting failures
and long-lived workloads. The observer is separately instrumented but
remains under the same research process. Timing is susceptible to
workstation noise; randomized blocks and raw observations expose some
variability but do not remove it.

\textbf{Statistical validity.} Fifty payload variants are not fifty
distinct failure mechanisms. Five repeated termination trials per cell
do not support a production risk estimate. Bootstrap intervals summarize
round-level latency variability on one host; no inference to all
machines or deployments is justified. Counts from targeted mutations and
bounded enumeration are not universal correctness measures.

\textbf{External validity.} No language model, autonomous planning loop,
prompt-injection benchmark, real vendor API, independent implementation,
regulated-firm pilot, production deployment, or comparative
developer-effort study is included. Real-network latency may dominate
local overhead, and distributed concurrency, storage failure, large
long-lived state, and policy complexity may change both correctness and
cost. Seven archetypes and scorecard thresholds remain unvalidated for
broad coverage and predictive value.

\textbf{Research provenance.} The repository and this study were
developed with AI assistance. Earlier maintainer reviews are historical
development context, not independent evidence; their reported 30
findings are not used as the primary empirical result here. The focused
reference audit preserves remaining verification issues rather than
claiming a complete literature or legal review.

\section{7 Conclusion}\label{conclusion}

OPCP provides an execution boundary in which proposed agent actions are
mediated by business-controlled contracts, authority, context
restrictions, and supplier rules. Scripted experiments characterize this
boundary: conservative handling of uncertainty avoids cross-supplier
duplication in the selected scenarios, persistence and supplier
idempotency address different recovery failures, routing and input size
affect local cost, and internally consistent evidence can still be
insufficient.

The conservative composition's equivalent fault behavior limits a
superiority claim. For AI-agent integration, the main result is an
explicit separation of action proposal, permission, effect, and
acknowledgement, together with the assumptions required to maintain it.
The study does not show that an LLM follows the response protocol or
that the resulting agent achieves safer or more useful tasks. Those
claims require end-to-end evaluation with real agents; migration-effort
and model-portability benefits require their own comparative evidence.

\section{Artifact availability}\label{artifact-availability}

The code base is publicly available as OPCP {[}26{]}. The accompanying
research artifact contains the frozen protocol, runners, raw CSV/JSON,
observer traces, source hashes, and figure-generation code in the paper
folder under \texttt{evaluation/v03/}. The research additions and
retained observations are supplied in the ancillary research archive.
The evaluated upstream implementation is pinned by its commit and
content hashes. The supplied LaTeX and vector figures reproduce this
paper.

\section{References}\small\sloppy\label{references}

\begin{enumerate}
\def\labelenumi{[\arabic{enumi}]}\setlength{\itemsep}{2pt}\setlength{\parsep}{0pt}
\item
  D. Waghmare, S. Sufi. \href{https://arxiv.org/abs/2611.22075v1}{Can
  You Undo It? Reversibility Classes for AI Agent Actions}.
  arXiv:2611.22075, 17 November 2026 (preprint).
  \href{https://github.com/TwoMuchLabs/reversibility-classes}{Project
  specification}.
\item
  \href{https://eur-lex.europa.eu/eli/reg/2022/2554/oj}{Regulation (EU)
  2022/2554 on digital operational resilience for the financial sector
  (DORA)}, 14 December 2022.
\item
  \href{https://www.legislation.gov.au/F2026L00475/asmade/2026-04-29/text/original/epub/OEBPS/document_1/document_1.html}{Prudential
  Standard CPS 230 Operational Risk Management}, Australian Prudential
  Regulation Authority.
\item
  \href{https://www.mas.gov.sg/publications/consultations/2026/consultation-paper-on-proposed-guidelines-on-third-party-risk-management}{Consultation
  Paper on Proposed Guidelines on Third-Party Risk Management
  (P004-2026)}, Monetary Authority of Singapore, 6 March 2026.
\item
  \href{https://www.esma.europa.eu/press-news/esma-news/european-supervisory-authorities-designate-critical-ict-third-party-providers}{European
  Supervisory Authorities designate critical ICT third-party providers},
  November 2025.
\item
  \href{https://www.bis.org/publications/202512-guidelines-principles-sound-management-third-party-risk.pdf}{Principles
  for the sound management of third-party risk}, Basel Committee on
  Banking Supervision, 10 December 2025.
\item
  \href{https://www.fsb.org/uploads/P14112024.pdf}{The Financial
  Stability Implications of Artificial Intelligence}, Financial
  Stability Board, 14 November 2024.
\item
  \href{https://www.bankingsupervision.europa.eu/press/supervisory-newsletters/newsletter/2025/html/ssm.nl250219_2.en.html}{Outsourcing
  trends in the banking sector}, ECB Banking Supervision, Supervision
  Newsletter, 19 February 2025.
\item
  \href{https://www.eiopa.europa.eu/publications/esas-joint-report-major-ict-related-incidents-2025_en}{Joint
  report on major ICT-related incidents in 2025 (JC 2026 16)}, EBA,
  EIOPA and ESMA, 3 June 2026.
\item
  C. Peukert.
  \href{https://doi.org/10.1080/0960085X.2018.1529374}{Determinants and
  heterogeneity of switching costs in IT outsourcing: estimates from
  firm-level data}. \emph{European Journal of Information Systems}
  28(3):291-317, 2019.
\item
  J. Farrell, P. Klemperer.
  \href{https://ideas.repec.org/h/eee/indchp/3-31.html}{Coordination and
  Lock-In: Competition with Switching Costs and Network Effects}.
  \emph{Handbook of Industrial Organization}, vol.~3, ch.~31, Elsevier,
  2007.
\item
  Y. Wang et al.~\href{https://arxiv.org/abs/2512.01420}{PromptBridge:
  Cross-Model Prompt Transfer for Large Language Models}.
  arXiv:2512.01420, 2025 (preprint).
\item
  Model Context Protocol.
  \href{https://github.com/modelcontextprotocol/modelcontextprotocol/blob/main/schema/2026-07-28/schema.json}{Tool
  schema, specification 2026-07-28}, \texttt{ToolAnnotations}. Accessed
  19 November 2026.
\item
  OpenTelemetry.
  \href{https://github.com/open-telemetry/semantic-conventions-genai}{Semantic
  conventions for generative AI}, read at commit \texttt{e6658f1}, 6
  November 2026.
\item
  Open Policy Agent.
  \href{https://www.openpolicyagent.org/docs}{Official documentation}.
  Accessed 19 November 2026.
\item
  H. Garcia-Molina, K. Salem.
  \href{https://doi.org/10.1145/38713.38742}{Sagas}. \emph{Proceedings
  of ACM SIGMOD}, pp.~249-259, 1987.
  \href{https://www.cs.princeton.edu/research/techreps/598}{Author
  technical-report record}.
\item
  OCEL. \href{https://www.ocel-standard.org/}{Object-Centric Event Log
  2.0}. Official specification site. Accessed 19 November 2026.
\item
  Object Management Group.
  \href{https://www.omg.org/spec/BPMN/2.0.2}{Business Process Model and
  Notation (BPMN), version 2.0.2}, January 2014.
\item
  BIAN. \href{https://github.com/bian-official/public}{Service
  Landscape, release 14.0.0}.
  \href{https://www.apqc.org/process-frameworks}{APQC Process
  Classification Framework}.
\item
  I. Ong et
  al.~\href{https://proceedings.iclr.cc/paper_files/paper/2025/hash/5503a7c69d48a2f86fc00b3dc09de686-Abstract-Conference.html}{RouteLLM:
  Learning to Route LLMs from Preference Data}. ICLR 2025.
  \href{https://arxiv.org/abs/2406.18665v4}{Author manuscript}.
\item
  FINOS. \href{https://air-governance-framework.finos.org/}{AI
  Governance Framework}, version 3 (read in part).
\item
  K. Tallam. \href{https://arxiv.org/html/2606.12320}{A Five-Plane
  Reference Architecture for Runtime Governance of Production AI
  Agents}. arXiv:2606.12320, 2026.
\item
  K. Mohsenzadegan, V. Tavakkoli, K. Kyamakya.
  \href{https://arxiv.org/abs/2609.37457}{VeriWeave Govern:
  Evidence-Gated Deterministic Runtime Governance for Enterprise AI
  Agents}. arXiv:2609.37457, 2026.
\item
  R. W. Sure. \href{https://arxiv.org/pdf/2607.03516}{AGL-1: The
  Enterprise AI Governance Layer as a Control Plane for Trusted
  Enterprise Intelligence}. arXiv:2607.03516, 2026.
\item
  M. Hsu, M. Bankhwal, B. Moissinac, K. Werbach, L. Szpruch, B.
  Hillenbrand. \href{https://arxiv.org/pdf/2610.04123}{Agent Reliability
  Profiles in Financial Services}. arXiv:2610.04123, 2026.
\item
  D. Waghmare, S. Sufi. \href{https://github.com/TwoMuchLabs/opcp}{Open
  Process Control Plane}, package version 0.3.2, 16 November 2026.
\item
  H. Zhang, A. Cardoza, P. B. Chen, S. Angel, V. Liu.
  \href{https://www.usenix.org/conference/osdi20/presentation/zhang-haoran}{Fault-tolerant
  and Transactional Stateful Serverless Workflows}. OSDI 2020,
  pp.~1187-1204.
  \href{https://www.cis.upenn.edu/~sga001/papers/beldi-osdi20.pdf}{Author
  PDF}.
\item
  Z. Jia, E. Witchel.
  \href{https://doi.org/10.1145/3477132.3483541}{Boki: Stateful
  Serverless Computing with Shared Logs}. SOSP 2021, pp.~691-707.
  \href{https://www.cs.utexas.edu/~witchel/pubs/jia21sosp-boki.pdf}{Author
  PDF}.
\item
  E. Debenedetti, J. Zhang, M. Balunovic, L. Beurer-Kellner, M. Fischer,
  F. Tramèr.
  \href{https://proceedings.neurips.cc/paper_files/paper/2024/hash/97091a5177d8dc64b1da8bf3e1f6fb54-Abstract-Datasets_and_Benchmarks_Track.html}{AgentDojo:
  A Dynamic Environment to Evaluate Prompt Injection Attacks and
  Defenses for LLM Agents}. \emph{Advances in Neural Information
  Processing Systems} 37, Datasets and Benchmarks Track, 2024.
\item
  Y. Ruan, H. Dong, A. Wang, S. Pitis, Y. Zhou, J. Ba, Y. Dubois, C. J.
  Maddison, T. Hashimoto.
  \href{https://arxiv.org/abs/2309.15817v2}{Identifying the Risks of LM
  Agents with an LM-Emulated Sandbox}. arXiv:2309.15817v2, 2024 (author
  manuscript).
\end{enumerate}

\end{document}